\documentclass[letterpaper]{article} % DO NOT CHANGE THIS

\usepackage{aaai2026}

\usepackage{times}  % DO NOT CHANGE THIS
\usepackage{helvet}  % DO NOT CHANGE THIS
\usepackage{courier}  % DO NOT CHANGE THIS
\usepackage[hyphens]{url}  % DO NOT CHANGE THIS
\usepackage{graphicx} % DO NOT CHANGE THIS
\usepackage{natbib}  % DO NOT CHANGE THIS AND DO NOT ADD ANY OPTIONS TO IT
\usepackage{caption} % DO NOT CHANGE THIS AND DO NOT ADD ANY OPTIONS TO IT
\usepackage{algorithm}
\usepackage{algorithmic}
\nocopyright

\usepackage{newfloat}
\usepackage{listings}
\DeclareCaptionStyle{ruled}{labelfont=normalfont,labelsep=colon,strut=off} % DO NOT CHANGE THIS
\floatstyle{ruled}
\newfloat{listing}{tb}{lst}{}
\floatname{listing}{Listing}
\usepackage{booktabs}
\usepackage{multirow}
\usepackage{textcomp} 
\usepackage{amsmath}

\title{Physics Knowledge in Frontier Models: A Diagnostic Study of Failure Modes}

\author {
    Ieva Bagdonaviciute\textsuperscript{\rm 1},
    Vibhav Vineet\textsuperscript{\rm 2},
}
\affiliations {
    \textsuperscript{\rm 1}University of Illinois Urbana–Champaign\\
    \textsuperscript{\rm 2}Microsoft Research\\
    ievab2@illinois.edu, vibhav.vineet@microsoft.com
}

\makeatletter
\def\@copyright{}\def\@copyrightspace{}
\makeatother

\begin{document}

\maketitle

%====================== ABSTRACT

\begin{abstract}
While recent Vision–Language Models (VLMs) have achieved impressive progress, it remains difficult to determine why they succeed or fail on complex reasoning tasks. Traditional benchmarks evaluate \emph{what} models can answer correctly, not \emph{why} they succeed or fail. In this work, we perform a failure-mode analysis of six frontier VLMs on three physics-based benchmarks -- Physion, Physion++, and CLEVRER -- by introducing custom subtests (for Physion and Physion++) and an integration of existing benchmark categories (for CLEVRER) to factor benchmark performance into distinct, testable capabilities. These subtests isolate \emph{perception} (object, color, and occlusion recognition) and \emph{physics understanding} (motion prediction and spatial reasoning), enabling us to test whether models attend to the correct entities and dynamics underlying their answers. Counterintuitively, subtest mastery correlates only weakly with benchmark accuracy: models often answer correctly without grounding in perception or physics. This suggests that current VLMs sometimes achieve benchmark scores for the wrong reasons, underscoring the need for diagnostics that expose hidden failure modes beyond aggregate metrics.
\end{abstract}


%====================== INTRODUCTION

\section{Introduction}

Recent progress in multimodal and embodied AI has produced models that can describe scenes and predict events from visual input. Yet benchmark performance alone offers limited insight into whether these models truly understand the underlying physical dynamics -- properties such as mass, friction, gravity, collisions, and occlusion. Human cognition tightly couples perception and reasoning: accurate perception enables better prediction and causal inference. In real-world settings, this integration supports safe decision-making and planning, making physical understanding essential for intelligent systems.

\begin{figure*}[t]
  \centering
  \includegraphics[width=0.75\linewidth]{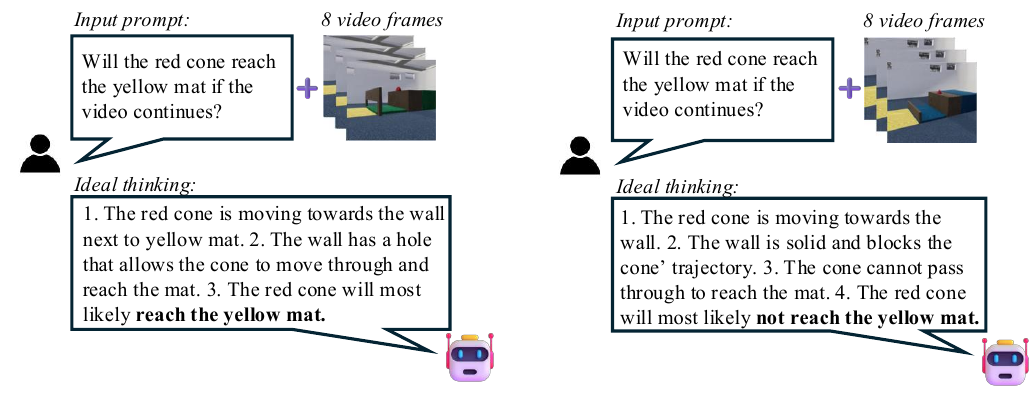}
  \caption{Illustrative examples of physical reasoning using Physion++. 
  Left: a cone approaches a wall with a hole and passes through to reach the goal. 
  Right: the same cone approaches a solid wall and is blocked.}
  \label{fig:examples}
\end{figure*}

Consider a video of a cone rolling toward a wall with a hole (Figure~\ref{fig:examples}, left). A model should infer that the cone passes through the opening to reach its goal; if the wall is solid, it should reason that the cone is blocked (Figure~\ref{fig:examples}, right). Solving both cases requires recognizing objects and geometry while reasoning about physical properties such as speed, mass, friction, and material interactions.

However, many existing benchmarks collapse perception and physics reasoning into a single accuracy score, obscuring whether errors stem from failures to \emph{see} or to \emph{understand}. We address this with a \emph{diagnostic evaluation framework} that decomposes benchmark questions into subtests for \emph{perception} (object, color, occlusion) and \emph{physics reasoning} (motion prediction, spatial relations), revealing whether models attend to the correct entities and interactions.

We apply this framework to six frontier VLMs across 
Physion~\cite{bear2022physionevaluatingphysicalprediction}, Physion++~\cite{tung2023physionpp}, and CLEVRER~\cite{yi2020clevrercollisioneventsvideo}. Counterintuitively, subtest mastery is only weakly correlated with benchmark accuracy. Models often answer benchmark questions correctly without detecting the relevant objects or motions, and conversely fail evaluation questions despite correct perceptual and physical grounding. These inconsistencies show that current VLMs can score well without integrating perception and physics into coherent causal representations. Our analysis reframes benchmark evaluation as failure-mode discovery, revealing how models rely on superficial cues rather than genuine physical reasoning.

%============= METHODOLOGY ===============

\section{Methodology}
\label{sec:methodology}

We evaluate six frontier VLMs (GPT-4o, GPT-O1, VideoLLaVA-7B, VideoLLaVA-NeXT-7B, Qwen2.5-VL-7B-Instruct, and InternVL-4B) across three physics-based reasoning benchmarks: Physion, Physion++ \cite{bear2022physionevaluatingphysicalprediction,tung2023physionpp}, and CLEVRER \cite{yi2020clevrercollisioneventsvideo}. Our goal is to disentangle whether model success on evaluation questions arises from accurate \textit{perception} and/or genuine \textit{physics reasoning}.

\subsection{Diagnostic Framework}

\begin{figure}[tb]
  \centering
  \includegraphics[width=\linewidth]{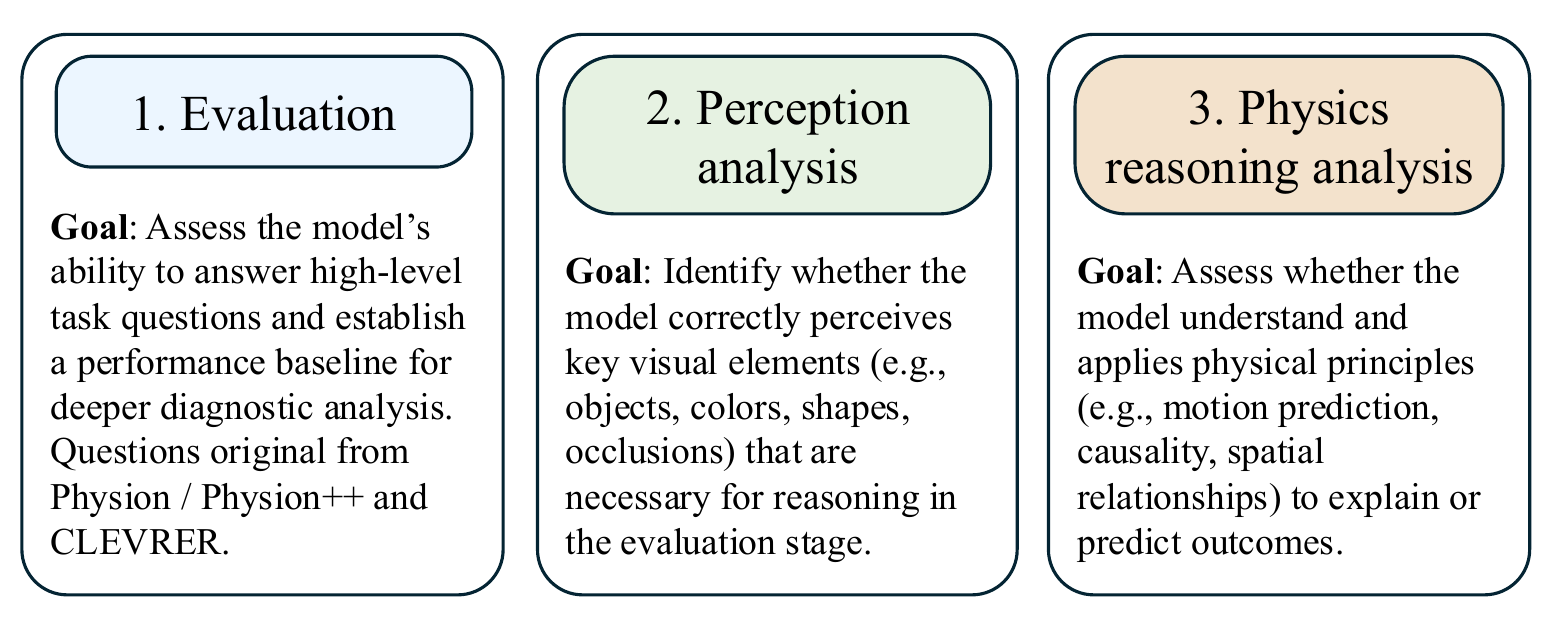}
  \caption{General pipeline for evaluation and analysis.}
  \label{fig:generalpipeline}
\end{figure}

Our diagnostic evaluation proceeds in two stages:  
(1) a standard benchmark evaluation measuring task accuracy, and  
(2) a diagnostic analysis measuring perception and physics subskills.  
This framework is summarized in Figure~\ref{fig:generalpipeline}.

In the first stage, models answer the original benchmark questions.
In the second and third stages, models are tested on additional subtests that probe whether they correctly perceive the key entities and interactions required for reasoning.  
We refer to these as \textit{perception} and \textit{physics} diagnostics.

\subsection{Physion and Physion++ Setup}
The Physion and Physion++ datasets test intuitive physics through video-based prediction of object motion and interaction outcomes. Each video depicts a target object (e.g., a red cone) interacting with other elements (e.g., walls, holes, ramps) while moving toward a goal (e.g., a yellow mat).  

We follow the two standard evaluation modes defined by \cite{tung2023physionpp}:  
\textit{with property} (we refer to as WP) videos are truncated, requiring models to predict the final outcome, while \textit{fully observed} (we refer to as WOP) videos show the full trajectory for verification.

% \paragraph{Evaluation.}
% The benchmark task asks whether the target object reaches the goal at the end of the video.

% \paragraph{Perception diagnostics.}
% We introduce three perception subtests that assess whether the model detects:  
% (1) the \textbf{target} (main moving object),  
% (2) the \textbf{goal} (destination), and  
% (3) the \textbf{latent object} (e.g., occluders or holes that affect motion), all of which necessary to see to see in order to correctly answer the evaluation question.

% \paragraph{Physics reasoning diagnostics.}
% We further probe physics understanding using two subtests:  
% (1) \textbf{motion prediction}, testing whether the model anticipates object trajectories, and  
% (2) \textbf{spatial relationships}, testing causal and spatial interactions.  
\textbf{1. Evaluation.} The benchmark task asks whether the target object reaches the goal at the end of the video. 
\textbf{2. Perception diagnostics.} We introduce three subtests that assess whether the model detects (1) the \emph{target} (main moving object), (2) the \emph{goal} (destination), and (3) the \emph{latent object} (e.g., occluders or holes that affect motion), all of which are necessary to correctly answer the evaluation question. 
\textbf{3. Physics reasoning diagnostics.} We further probe physics understanding using two subtests: (1) \emph{motion prediction}, testing whether the model anticipates object trajectories, and (2) \emph{spatial relationships}, testing causal and spatial interactions. 

A full illustration of our Physion/Physion++ diagnostic structure and example prompts is shown in Figure~\ref{fig:pipelines}~(top).

% \begin{figure*}[t]
%   \centering
%   \includegraphics[width=0.8\linewidth]{Physion_setup.png}
%   \caption{\textbf{Evaluation and analysis pipeline used for Physion and Physion++ datasets.}
%   Diagnostic subtests isolate evaluation, perception (target, goal, latent), and physics (motion, spatial) reasoning.}
%   \label{fig:Physionpipeline}
% \end{figure*}

\subsection{CLEVRER Setup}

% The CLEVRER dataset consists of synthetic object-collision videos with four categories of questions:  
% \textit{descriptive}, \textit{explanatory}, \textit{predictive}, and \textit{counterfactual}.  
% We align these categories with our diagnostic framework:  
% counterfactual → evaluation,  
% descriptive + explanatory → perception,  
% and predictive → physics reasoning.  
% This alignment reflects CLEVRER’s original design: counterfactuals require causal reasoning over full scene dynamics, descriptive and explanatory questions probe surface-level recognition and causal attribution (analogous to perception), and predictive questions demand temporal and physical inference, making them a natural proxy for physics reasoning.

% \paragraph{Evaluation.}
% Counterfactual questions serve as the evaluation metric.

% \paragraph{Perception diagnostics.}
% Descriptive and explanatory questions probe object recognition and causal explanations of collisions.  

% \paragraph{Physics reasoning diagnostics.}
% Predictive questions test whether the model can anticipate future events.  

The CLEVRER dataset consists of synthetic object-collision videos featuring four categories of questions: \textit{descriptive}, \textit{explanatory}, \textit{predictive}, and \textit{counterfactual}.
Each question type probes a distinct level of understanding -- from basic perception to causal reasoning -- making CLEVRER a natural fit for our diagnostic framework. Descriptive and explanatory questions emphasize recognition and attribution of observed events, predictive questions assess temporal and physical inference, and counterfactual questions require reasoning about alternative outcomes.
Guided by these distinctions, we categorize the CLEVRER tasks as follows. 
\textbf{1. Evaluation.} Counterfactual questions serve as the primary evaluation metric, requiring reasoning over full scene dynamics and hypothetical outcomes. \textbf{2. Perception diagnostics.} Descriptive and explanatory questions probe object recognition and causal attribution within observed collisions. \textbf{3. Physics reasoning diagnostics.} Predictive questions test the model’s ability to anticipate future physical events. 

Appendix~\ref{sec:appendix} (Figure~\ref{fig:pipelines}, bottom) summarizes the diagnostic structure and example prompts for CLEVRER.

% \begin{figure*}[t]
%   \centering
%   \includegraphics[width=0.60\linewidth]{CLEVRER_setup.png}
%   \caption{\textbf{Evaluation and analysis pipeline used for CLEVRER dataset.}
%   Counterfactual questions measure evaluation accuracy; descriptive, explanatory, and predictive categories test perception and physics reasoning.}
%   \label{fig:CLEVRERpipeline}
% \end{figure*}

\subsection{Implementation Details}

All experiments used our custom pipeline built on the \texttt{EUREKA} framework \cite{eureka2024,eureka2025}, which supports modular dataset integration.  
Models were independently queried on each diagnostic question using ground-truth binary or multiple-choice answers.  
For Physion/Physion++, we analyze 2{,}110 videos with full diagnostic coverage, and for CLEVRER, validation videos with one question per category to ensure consistency across models.

%====================== RESULTS
\section{Results}

\subsection{Part I: Evaluation and Diagnostic Subtests}

\begin{figure*}[t]
  \centering
  \includegraphics[width=0.8\linewidth]{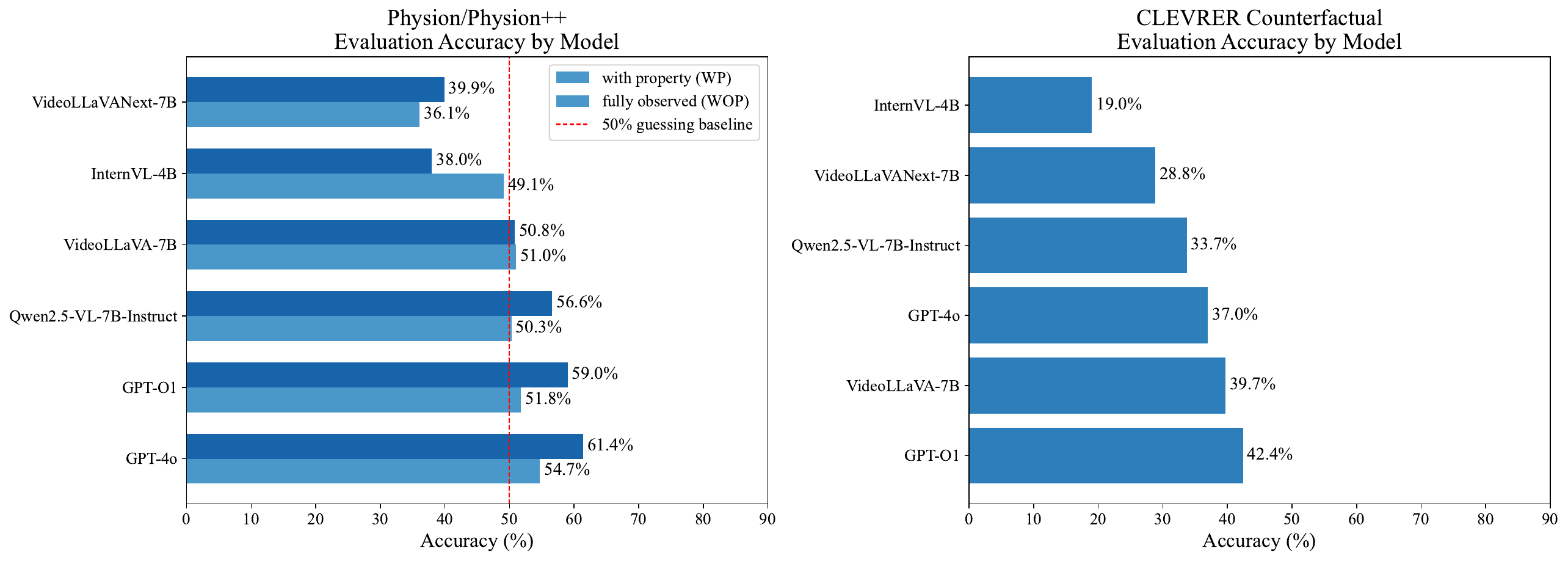}
  \caption{\textbf{Evaluation accuracy.} Left: Physion/Physion++ under WP and WOP. Right: CLEVRER counterfactual.}
  \label{fig:evaluation}
\end{figure*}

\paragraph{Evaluation.}
As shown in Figure~\ref{fig:evaluation}, frontier VLMs struggle with physical reasoning: overall accuracy on Physion/Physion++ and CLEVRER counterfactuals remains modest, but a few models average above chance across different settings. 

\begin{figure*}[t]
  \centering
  \includegraphics[width=0.8\linewidth]{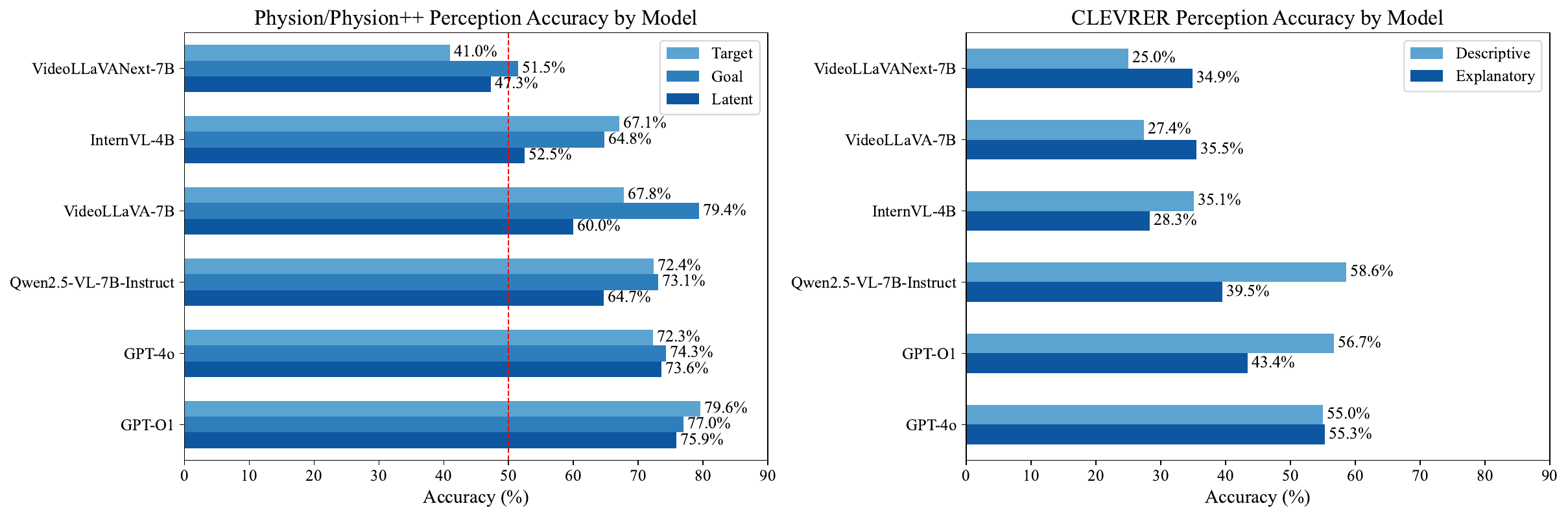}
  \caption{\textbf{Perception accuracy.} Left: Physion/Physion++ (target, goal, latent). Right: CLEVRER (descriptive, explanatory).}
  \label{fig:perception_composite}
\end{figure*}

\begin{figure*}[t]
  \centering
  \includegraphics[width=0.8\linewidth]{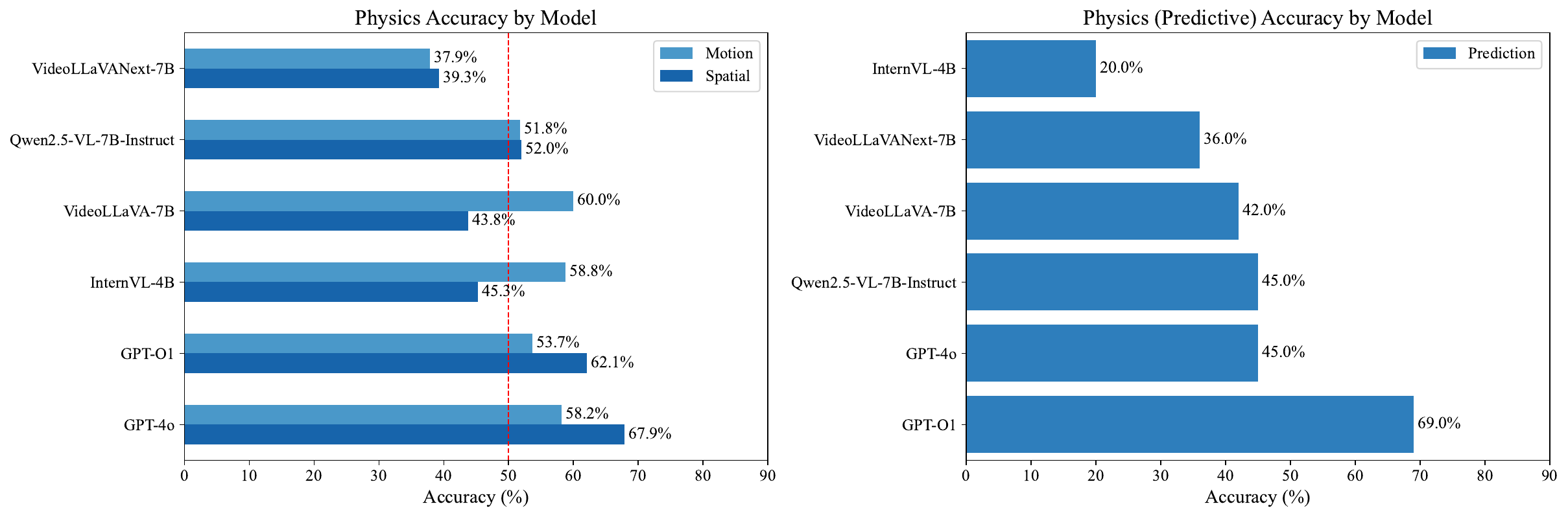}
  \caption{\textbf{Physics reasoning performance.} Left: Physion/Physion++ (motion, spatial). Right: CLEVRER (predictive).}
  \label{fig:physics_composite}
\end{figure*}

\paragraph{Diagnostics.}
Perception subtests (targets/goals/latent; descriptive/explanatory) are stronger than physics subtests (motion/spatial; predictive) for most models \mbox{(Figs.~\ref{fig:perception_composite}–\ref{fig:physics_composite})}. Perception is stronger overall, though some models lag in certain context, and physics diagnostics show clear gaps in dynamic understanding.

% =========================================================
% PART II — FAILURE-MODE ANALYSIS (TABLES)
% =========================================================
\subsection{Part II: Failure-Mode Analysis}

While some models appear to perform poorly across both evaluation and diagnostic tasks, the underlying errors often stem from different sources, indicating that benchmark accuracy alone offers an incomplete view of model competence.

\paragraph{Physion/Physion++.}
  
Intuitively, models passing all diagnostic subtests should exhibit higher evaluation accuracy (\(P(\text{correct}|\text{passed}) \gg P(\text{correct}|\text{failed})\)), with most correct evaluations occurring when all subtests are passed (\(k=5\)). However, Tables~\ref{tab:physionblockdelta}--\ref{tab:physionsuccessbins} show the opposite: \text{$\Delta$ Acc.} are small or even negative, and most correct answers occur with only three or four subtests passed. This suggests models often succeed without full perceptual or physical grounding, relying on shortcut reasoning rather than true understanding.

\begin{table}[!t]
\centering
\scriptsize
\setlength{\tabcolsep}{4pt}
\renewcommand{\arraystretch}{1.05}

\begin{tabular}{lcccc}
\toprule
\textbf{Model} & \textbf{Block} &
\textbf{P(correct$|$passed)} &
\textbf{P(correct$|$failed)} &
\textbf{$\Delta$ Acc.} \\
\midrule
GPT-4o & perc. & 56.0\% & 54.4\% & 1.5\% \\
       & phys. & 56.3\% & 54.4\% & 2.0\% \\
GPT-o1 & perc. & 54.1\% & 50.8\% & 3.2\% \\
       & phys. & 60.1\% & 48.6\% & 11.5\% \\
VideoLLaVA-7B & perc. & 49.7\% & 52.2\% & -2.5\% \\
               & phys. & 58.0\% & 49.2\% & 8.8\% \\
VideoLLaVA-NeXT-7B & perc. & 39.0\% & 36.5\% & 2.5\% \\
                    & phys. & 38.4\% & 36.5\% & 1.9\% \\
Qwen2.5-VL-7B-Inst. & perc. & 52.9\% & 49.5\% & 3.4\% \\
                     & phys. & 48.0\% & 51.7\% & -3.7\% \\
InternVL-4B & perc. & 46.9\% & 49.2\% & -2.4\% \\
            & phys. & 47.8\% & 48.9\% & -1.1\% \\
\bottomrule
\end{tabular}

\caption{\textbf{Physion/Physion++ -- Block “pass” vs.\ evaluation accuracy (WP).}
Evaluation accuracy when a model passes all diagnostic subtests in a block (\(P(\text{correct}|\text{passed})\)) vs.\ when it fails at least one (\(P(\text{correct}|\text{failed})\)); $\Delta$ is the difference in percentage points.}
\label{tab:physionblockdelta}
\end{table}

\begin{table}[!t]
\centering
\scriptsize        % smaller font (fits one column cleanly)
\setlength{\tabcolsep}{5pt}
\renewcommand{\arraystretch}{1.1}

\begin{tabular}{lcccccc}
\toprule
\textbf{Model} &
\multicolumn{6}{c}{\textbf{Exactly $k$ diagnostic subtests passed}} \\
\cmidrule(lr){2-7}
 & \textbf{$k=5$} & \textbf{$k=4$} & \textbf{$k=3$} & \textbf{$k=2$} & \textbf{$k=1$} & \textbf{$k=0$} \\
\midrule
GPT-4o              & 20.6\% & 36.4\% & 24.2\% & 13.6\% & 4.7\% & 0.4\% \\
GPT-o1              & 18.2\% & 38.0\% & 29.0\% & 12.2\% & 2.3\% & 0.4\% \\
VideoLLaVA-7B       & 7.9\%  & 25.8\% & 35.5\% & 26.2\% & 4.2\% & 0.5\% \\
VideoLLaVA-NeXT-7B  & 1.8\%  & 9.9\%  & 26.1\% & 33.4\% & 25.3\% & 3.5\% \\
Qwen2.5-VL-7B-Instruct & 9.3\%  & 29.7\% & 35.2\% & 17.8\% & 7.3\% & 0.7\% \\
InternVL-4B         & 6.4\%  & 22.9\% & 33.4\% & 25.0\% & 10.8\% & 1.4\% \\
\bottomrule
\end{tabular}

\caption{\textbf{Physion/Physion++ -- Diagnostic subtest breakdown.}
Distribution of diagnostic subtests passed (\(k=0\)–\(5\)) on videos where the model’s evaluation was correct.
}
\label{tab:physionsuccessbins}
\end{table}

\paragraph{CLEVRER: conditioning on counterfactual (evaluation).}
Tables~\ref{tab:clevrerdesctable}--\ref{tab:clevrerpredtable} report descriptive, explanatory, and predictive accuracy conditioned on the counterfactual bin (\( \text{CF}=100\%,50\%,0\% \)). Trends are non-monotonic across models, reinforcing that stronger perception or predictive skill does not reliably imply counterfactual success. In fact, oftentimes models maintain stable or even improved accuracy when counterfactual overlap decreases, suggesting that performance on observable (descriptive) or causal (explanatory/predictive) questions does not translate to reasoning about unseen alternatives -- despite its cognitive dependence on first correctly perceiving and understanding the scene.

% Tables~\ref{tab:clevrerdesctable}--\ref{tab:clevrerpredtable} report descriptive, explanatory, and predictive accuracy conditioned on the counterfactual bin (\( \text{CF}=100\%,50\%,0\% \)). Trends are non-monotonic across models, showing that stronger perceptual or predictive skill does not imply counterfactual success. Some models (e.g., GPT-4o, Qwen2.5) even maintain or improve accuracy as counterfactual overlap decreases, suggesting that performance on observable (descriptive) or causal (explanatory/predictive) questions does not extend to reasoning about unseen alternatives. This misalignment indicates that passing perception and understanding tests alone is insufficient for genuine counterfactual reasoning---despite its dependence on correctly perceiving and understanding the scene.

\begin{table}[!t]
\centering
\scriptsize
\setlength{\tabcolsep}{6pt}
\renewcommand{\arraystretch}{1.1}

\begin{tabular}{lccc}
\toprule
\textbf{Model} &
\multicolumn{3}{c}{\textbf{Descriptive accuracy when}} \\
\cmidrule(lr){2-4}
 & \textbf{CF = 100\%} & \textbf{CF = 50\%} & \textbf{CF = 0\%} \\
\midrule
GPT-4o                 & 59.1\% & 54.3\% & 54.1\% \\
GPT-o1                 & 56.3\% & 55.8\% & 58.0\% \\
VideoLLaVA-7B          & 28.3\% & 27.5\% & 26.8\% \\
VideoLLaVA-NeXT-7B     & 26.4\% & 22.3\% & 26.4\% \\
Qwen2.5-VL-7B-Instruct & 58.6\% & 58.6\% & 58.7\% \\
InternVL-4B            & 34.3\% & 35.5\% & 35.0\% \\
\bottomrule
\end{tabular}

\caption{\textbf{CLEVRER -- Descriptive vs.\ Counterfactual (CF) Bins.}
Average descriptive accuracy for CF conditions.}
\label{tab:clevrerdesctable}
\end{table}

\begin{table}[!t]
\centering
\scriptsize
\setlength{\tabcolsep}{6pt}
\renewcommand{\arraystretch}{1.1}

\begin{tabular}{lccc}
\toprule
\textbf{Model} &
\multicolumn{3}{c}{\textbf{Explanatory accuracy when}} \\
\cmidrule(lr){2-4}
 & \textbf{CF = 100\%} & \textbf{CF = 50\%} & \textbf{CF = 0\%} \\
\midrule
GPT-4o              & 53.1\% & 56.0\% & 57.1\% \\
GPT-o1              & 38.1\% & 53.6\% & 40.5\% \\
VideoLLaVA-7B       & 29.4\% & 33.7\% & 35.0\% \\
VideoLLaVA-NeXT-7B  & 18.2\% & 35.7\% & 35.2\% \\
Qwen2.5-VL-7B-Instruct & 33.3\% & 44.1\% & 43.8\% \\
InternVL-4B         & 5.6\%  & 25.0\% & 32.6\% \\
\bottomrule
\end{tabular}

\caption{\textbf{CLEVRER -- Explanatory vs.\ Counterfactual (CF) Bins.}
Average explanatory accuracy for CF conditions.}
\label{tab:clevrerexpltable}
\end{table}

\begin{table}[!t]
\centering
\scriptsize
\setlength{\tabcolsep}{6pt}
\renewcommand{\arraystretch}{1.1}

\begin{tabular}{lccc}
\toprule
\textbf{Model} &
\multicolumn{3}{c}{\textbf{Predictive accuracy when}} \\
\cmidrule(lr){2-4}
 & \textbf{CF = 100\%} & \textbf{CF = 50\%} & \textbf{CF = 0\%} \\
\midrule
GPT-4o              & 37.5\% & 50.0\% & 42.9\% \\
GPT-o1              & 57.1\% & 81.0\% & 62.2\% \\
VideoLLaVA-7B       & 52.9\% & 37.2\% & 42.5\% \\
VideoLLaVA-NeXT-7B  & 27.3\% & 28.6\% & 42.6\% \\
Qwen2.5-VL-7B-Instruct & 61.1\% & 52.9\% & 33.3\% \\
InternVL-4B         & 55.6\% & 22.7\% & 14.5\% \\
\bottomrule
\end{tabular}

\caption{\textbf{CLEVRER -- Predictive vs.\ Counterfactual (CF) Bins.}
Average predictive accuracy for CF conditions.}
\label{tab:clevrerpredtable}
\end{table}

%====================== DISCUSSION

\section{Discussion}

% Our findings reinforce long-standing critiques that current multimodal models act more like sophisticated pattern matchers than causal reasoners \cite{lake2017building,marcus2018deep,HOLZINGER2023100788}.  
% Rather than forming a coherent world model, VLMs use perception and physics as independent shortcuts: they can recognize surface features or pass isolated probes, yet these skills rarely yield reliable predictive or counterfactual reasoning. Many evaluation wins resemble statistical guesswork: correct answers without fully seeing the scene or reasoning about its dynamics, exposing a gap between pattern matching and causal inference that resembles hallucination.

Our findings reinforce long-standing critiques that current multimodal models often behave as sophisticated pattern matchers rather than causal reasoners \cite{lake2017building,marcus2018deep,HOLZINGER2023100788}.  
Instead of forming coherent world models, VLMs tend to treat perception and physics as separate shortcuts -- recognizing surface features without integrating them into consistent causal representations. This behavior exposes a persistent gap between pattern recognition and genuine physical reasoning.

%====================== CONCLUSION

\section{Conclusion}

We conducted a comprehensive failure-mode analysis of six frontier VLMs across three physics benchmarks -- CLEVRER, Physion, and Physion++ -- using a diagnostic framework that separates \emph{perception} (scene grounding) from \emph{physics reasoning} (temporal/spatial dynamics) and \emph{task evaluation}. Across datasets, benchmark accuracy is only weakly aligned with diagnostic competence. These findings emphasize the need for more granular, structured benchmarks that decompose evaluation into interconnected subtasks -- separating perception, causal inference, and counterfactual reasoning.  
Such diagnostics do not replace existing evaluations but complement them, providing a clearer lens on where models succeed, where they fail, and what forms of reasoning remain unsolved.

%======================

\section{Acknowledgments}

We thank the entire AI Frontiers Laboratory for their guidance and support throughout this project, with special gratitude to Vidhisha Balachandran and Besmira Nushi for their invaluable feedback and mentorship.

% \appendix
% \section{Appendix: Dataset Pipelines}

% This appendix summarizes the diagnostic pipelines and example questions used in our analyses for Physion/Physion++ and CLEVRER.

% \begin{figure*}[t]
%   \centering
%   % Top panel
%   \includegraphics[width=0.85\linewidth]{Physion_setup.png}

%   % Bottom panel
%   \includegraphics[width=0.57\linewidth]{CLEVRER_setup.png}

%   \caption{\textbf{Diagnostic pipelines.} \textit{Top:} Physion and Physion++—evaluation plus subtests isolating perception (target, goal, latent) and physics (motion, spatial) reasoning. \textit{Bottom:} CLEVRER—counterfactual questions for evaluation; descriptive and explanatory (perception) and predictive (physics) categories for diagnostics.}
%   \label{fig:pipelines}
% \end{figure*}

\appendix
\label{sec:appendix}

% \section{Appendix: Dataset Pipelines}

% This appendix summarizes the diagnostic pipelines and example questions used in our analyses for Physion/Physion++ and CLEVRER.

\begin{figure*}[t] % bottom, spans both columns
  \centering
  % Top panel
  \includegraphics[width=0.83\linewidth]{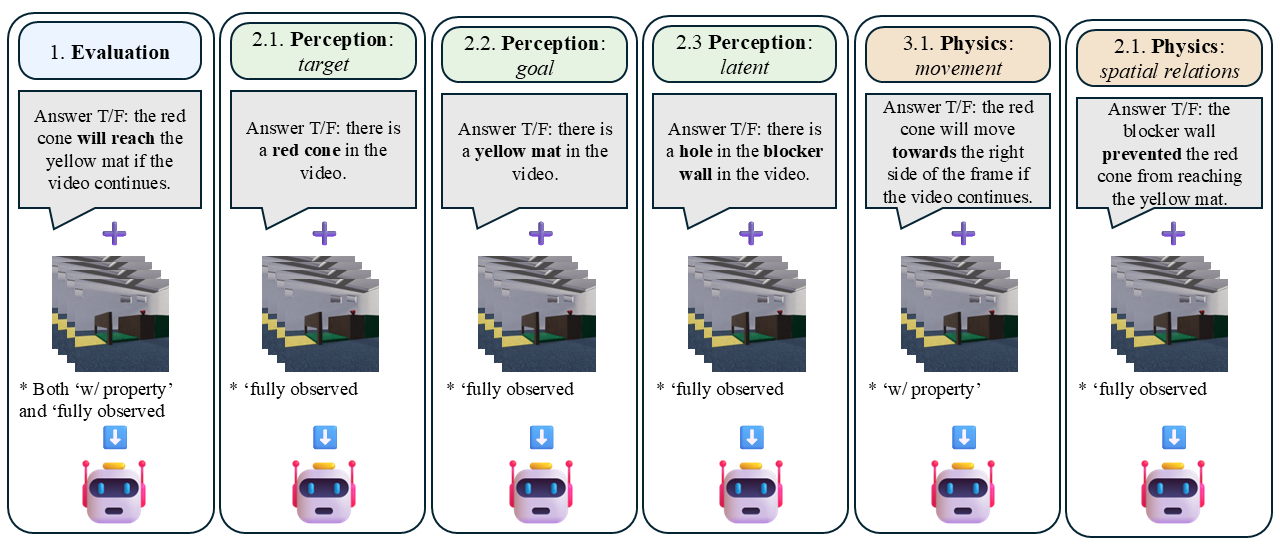}\par\vspace{0.5ex}
  % Bottom panel
  \includegraphics[width=0.55\linewidth]{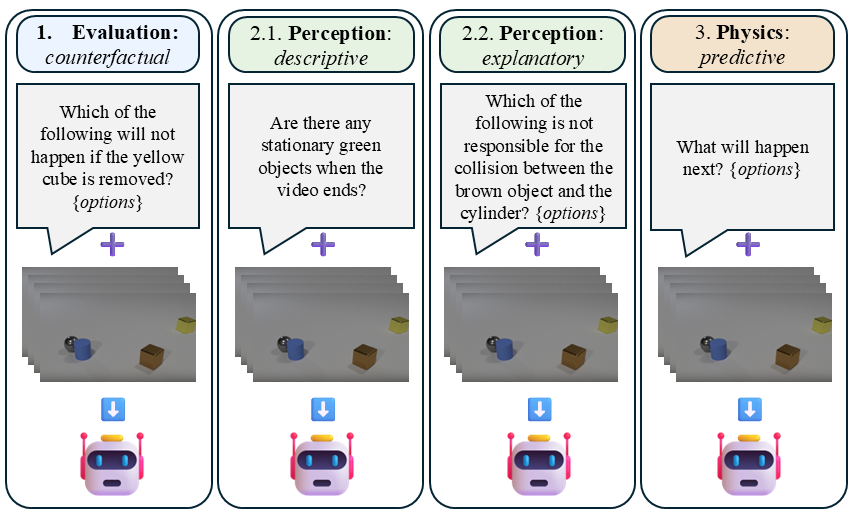}
  \caption{\textbf{Diagnostic pipelines.} \textit{Top:} Physion and Physion++ -- evaluation plus subtests isolating perception (target, goal, latent) and physics (motion, spatial) reasoning. \textit{Bottom:} CLEVRER -- counterfactual questions for evaluation; descriptive and explanatory (perception) and predictive (physics) categories for diagnostics.}
  \label{fig:pipelines}
\end{figure*}

\section{Appendix: Dataset Pipelines}

Figure~\ref{fig:pipelines} summarizes the diagnostic setups and example questions for Physion/Physion++ and CLEVRER used in our analyses.

%========== REFERENCES =======

%\bibliographystyle{aaai}
\bibliography{references}

\end{document}